\documentclass[11pt]{article}
\pdfoutput=1
\usepackage[final]{coling}

\usepackage{etoolbox}
\usepackage{times}
\usepackage{latexsym}

\usepackage[T1]{fontenc}

\usepackage[utf8]{inputenc}

\usepackage{microtype}

\usepackage{inconsolata}

\usepackage{graphicx}

\title{Development of Pre-Trained Transformer-based Models for the Nepali Language}

\author{
  \textbf{Prajwal Thapa\textsuperscript{*}},
  \textbf{Jinu Nyachhyon\textsuperscript{*}},
  \textbf{Mridul Sharma\textsuperscript{*}},
  \textbf{Bal Krishna Bal}
\\
\\
  \ Information and Language Processing Research Lab (ILPRL), Kathmandu University,
\\
  \small{
    Corresponding Author: \href{mailto:prazzwalthapa87@gmail.com}{prazzwalthapa87@gmail.com}
  }
}

\begin{document}
\maketitle
\begin{abstract}
Transformer-based pre-trained language models have dominated the field of Natural Language Processing (NLP) for quite some time now. However, the Nepali language, spoken by approximately 32 million people worldwide, remains significantly underrepresented in this domain. This underrepresentation is primarily attributed to the scarcity of monolingual data corpora and limited available resources for the Nepali language. While existing efforts have predominantly concentrated on basic encoder-based models, there is a notable gap in the exploration of decoder-based architectures. To address this gap, we have collected 27.5 GB of Nepali text data, approximately 2.4x larger than any previously available Nepali language corpus. Leveraging this data, we pre-trained three different models i.e., BERT, RoBERTa, and GPT-2, exclusively for the Nepali Language. Furthermore, we performed instruction tuning and explored its potential for monolingual Nepali data, providing a foundation for future research. Our models outperformed the existing best model by 2 points on Nep-gLUE benchmark, scoring 95.60 and also outperformed existing models on text generation tasks, demonstrating improvements in both understanding and generating Nepali text.

\end{abstract}

\section{Introduction}

In recent years, Natural Language Processing (NLP) has undergone a remarkable evolution, transitioning from traditional rule-based to statistical methods to sophisticated deep learning architectures. The initial approaches, such as n-grams \cite{Goodman2001} and rule-based systems, laid the groundwork for understanding language. But these methods faced significant limitations in handling the complexities of natural language and general human communication, which often involves subtle nuances, contextual dependencies, and varying linguistic structures. 

The introduction of Recurrent Neural Networks (RNNs) \cite{Mikolov2010} and Long Short-Term Memory networks (LSTMs) \cite{Sundermeyer2012} for language modeling marked a significant advancement, allowing models to process sequential data more effectively. RNNs and LSTMs brought notable improvements in tasks like language modeling and sequence prediction. However, they still encountered challenges with long-range dependencies and computational efficiency, which limited their scalability and performance. These models require substantial computational resources and face difficulties in maintaining consistent performance across varying lengths of text and contexts. The development of the self-attention mechanism \cite{Vaswani2017} marked a pivotal moment in NLP, enabling models to capture dependencies in text more effectively. The self-attention mechanism, integral to the Transformer architecture, allows models to weigh the importance of different words in a sequence, facilitating a more nuanced understanding of context. This was further enhanced by the concept of self-supervised model pre-training, where models like ELMo \cite{Peters2018}, BERT \cite{Devlin2019} and GPT \cite{Radford2018}, leveraged vast amounts of unlabeled text data to learn general language representations. These models demonstrated unprecedented performance improvements across a range of NLP tasks.

Further advancements in NLP include instruction tuning \cite{Wei2021, Wang2022}, which trains models on instruction-output pairs to improve their ability to follow user commands.This method enhances the model’s ability to follow specific user commands and adapt to diverse application scenarios. Instruction tuning has proven effective in improving the versatility and responsiveness of models, allowing them to better handle varied tasks and user interactions. Models, through unsupervised pre-training on large corpora, and instruction-tuning have demonstrated the ability to generalize across various tasks, achieving state-of-the-art results.

Nepali is spoken by over 32 million people worldwide. Syntactically, the Nepali language differs significantly from English. In English, the typical sentence structure follows a Subject-Verb-Object (SVO) order whereas Nepali employs a Subject-Object-Verb (SOV) structure \cite{Timilsina2022}. Nepali language incorporates a complex system of noun, adjective and verb inﬂections. Nouns have a system of gender, case and number \cite{Bal2004}. This fundamental difference in syntactic arrangement highlights the unique characteristics of Nepali and underscores the challenges pertinent to natural language processing tasks in the Nepali language.

Our motivation for developing a monolingual language model for Nepali language comes from recent advancements in natural language processing, particularly the success of large-scale pre-trained models. However, the majority of these developments have focused on high-resource languages, leaving a gap in the availability of robust models for low-resource languages like Nepali. To address this disparity, we developed pre-trained monolingual language models for the Nepali language. Our first steps included compiling a dataset, large enough to develop pre-trained language models. We assembled approximately 27.5 GB of text data by scraping the top 99 Nepali news websites, representing the largest Nepali language dataset to date. We also used an instruction-tuning dataset to explore the potential of instruction-tuned models for Nepali, providing a foundation for future advancements. 

This paper outlines our methods for developing pre-trained models and presents a thorough evaluation of their performance and comparison with existing models, setting new standards for Nepali NLP and contributing significantly to research on low-resource languages.

\section{Related Works}

The development of pre-trained language models has been foundational in advancing NLP, and a variety of approaches have emerged over the years. In this section, we briefly review the key methods that have influenced the creation and evolution of language models.

\subsection{Unsupervised Pre-training Approaches}
Early methods for developing pre-training language models utilized unsupervised learning to create generalized representations, with word embeddings like Word2Vec \cite{Mikolov2013} and GloVe \cite{Pennington2014} establishing the foundation by learning vector representations of words from large text corpora. These embeddings enhanced performance across various NLP tasks by capturing semantic relationships in a continuous vector space. The advent of contextualized word embeddings marked a significant advancement, exemplified by \cite{Peters2018}, which generated dynamic embeddings based on surrounding text using bidirectional LSTM networks, leading to improved results in benchmarks such as Question Answering and Named Entity Recognition.
The introduction of the Transformer architecture \cite{Vaswani2017} further revolutionized the field, giving rise to models like BERT \cite{Devlin2019}, RoBERTa \cite{Liu2019}, ELECTRA \cite{Clark2020}, DeBERTa \cite{He2020}  and GPT \cite{Radford2018}. BERT and other encoder model’s bidirectional training approach allowed it to predict masked words by considering both left and right context, achieving state-of-the-art results across numerous NLP tasks, while GPT's autoregressive method excelled in text generation and completion.

\subsection{Multilingual and Monolingual Language Models}
The release of multilingual models, such as m-BERT \cite{Pires2019}, XLM \cite{Lample2019} and XLM RoBERTa \cite{Conneau2020} which includes support for languages like Nepali, has further expanded the accessibility of NLP tools across different languages. While these models offer impressive results for many languages, their performance for languages other than English is still not up to the mark due to limited training data, issues with tokenization, lack of adequate vocabulary, and absence of techniques to handle linguistic diversity. 

Recently, numerous powerful monolingual models for languages other than English have emerged showing promising results such as NorBERT \cite{Kutuzov2021} for Nordic languages, FinBERT \cite{Virtanen2019} for Finnish language, HerBERT \cite{Mroczkowski2021} for Polish language, GBERT \cite{Chan2020} for German language, Chinese BERT \cite{Cui2021} for Chinese language, NepBERTa \cite{Timilsina2022} for Nepali language etc.  These models have demonstrated that optimizing tokenizers and architectures for specific languages can lead to substantial improvements in performance.

Few of the models have also focused on the Nepali language. IndicBERT \cite{Doddapaneni2022}  focused on several Indic languages, including Nepali, and demonstrated that language-specific models could outperform their multilingual counterparts on specialized tasks. NepBERTa \cite{Timilsina2022} introduced a BERT-based language model specifically for the Nepali language, trained on the largest monolingual Nepali corpus with 0.8 billion words collected from various sources. They also established the first Nepali Language Understanding Evaluation benchmark (Nep-gLUE). Similarly, NepaliBERT \cite{Pudasaini2023} also developed a monolingual BERT model specifically for the Nepali language. These models demonstrate the importance of optimizing tokenizers and architectures for addressing the unique characteristics of individual languages, especially those with complex syntactic and morphological structures like Nepali.

\subsection{Instruction Tuning on Low-Resource Language Models}
Instruction tuning has recently gained attention as a technique to substantially improve zero-shot performance on unseen tasks \cite{Wei2021, Wang2022, Ouyang2022}. This method involves fine-tuning pre-trained models on tasks that require the model to understand and execute explicit instructions, thereby increasing its adaptability and effectiveness across a variety of tasks. While this approach has been widely explored for high-resource languages, its potential in low-resource languages, such as Nepali, remains unexplored.

\section{Dataset}
This section describes the dataset used in our study, focusing on the methodologies implemented for data collection and preprocessing. Given the necessity for extensive training data in transformer-based language models, we compiled a dataset that is by far the largest one for the Nepali language.

\subsection{Dataset Collection}
Recently, the rise in digital content in Nepali has led to the increasing number of Nepali-language websites. This has opened doors to creating a comprehensive Nepali language corpus for which we performed web scraping across 99 Nepali news websites. As a result, we were able to gather a dataset totaling 30.4 GB of text data, which is significantly larger than existing resources.

We made a deliberate decision not to include existing datasets, such as the Nepali Wikipedia \cite{Arora2020} dataset which is less than 1GB, the OSCAR dataset \cite{OrtizSuarez2019} which is approximately 3GB, and the 12.5 GB dataset from NepBERTa \cite{Timilsina2022}. This choice was based on the fact that all these existing datasets were also scraped from news websites, which overlapped with our sources. To avoid duplication and ensure the uniqueness of our dataset, we opted to scrape all content from scratch. 

For Instruction Tuning, we utilized the publicly available Nepali alpaca dataset \cite{Kafley2024}) containing 52k rows of instructions. We cleaned the data by removing/translating all the non-Nepali texts. After the cleaning process, we achieved 40k rows of instructions.

\subsection{Dataset Preprocessing}
Following the data collection phase, we implemented a preprocessing pipeline to enhance the quality of the dataset. First, we implemented a deduplication process to remove redundant content, which was essential due to the extensive nature of our data collection. To address the challenge of multilingual content often found on news websites, we removed or translated the languages other than Nepali depending upon the context. We also developed specialized scripts to remove noise, eliminating non-textual elements such as HTML tags, special characters, and formatting artifacts common in web-scraped data. Additionally, text normalization techniques, including Unicode normalization, were applied to standardize character representation and maintain consistency. After these preprocessing steps, the dataset was refined and reduced to 27.5 GB, ensuring it was clean and well-suited for model training.

\subsection{Tokenization}
Tokenization is a crucial preprocessing step in natural language processing that breaks text into smaller units, such as words or subwords, enabling effective processing by language models. Traditional word piece tokenization methods \cite{Wu2016} rely on splitting text based on spaces and punctuation and often encounter limitations with out-of-vocabulary (OOV) words and morphological variations, leading to potential information loss. In contrast, Byte-Pair Encoding (BPE) \cite{Sennrich2016} tokenization addresses these limitations by segmenting words into subword units. BPE improves the handling of rare or unseen words by breaking them down into more manageable subword units, which helps in retaining meaningful information and maintaining consistency across different word forms. This method provides optimal balance between vocabulary size and coverage, which is particularly beneficial for morphologically rich languages like Nepali, where word forms can vary significantly.

For our study, we utilized the entire dataset to create two different BPE tokenizers, one with a vocabulary size of 30,522 and another with a vocabulary size of 50,256. These tokenizers were designed to optimize the balance between computational efficiency and linguistic coverage, ensuring that our language models could effectively process and understand Nepali text.

\section{BERT \& RoBERTa}
BERT and RoBERTa are both built upon the transformer encoder architecture, which serves as the foundation for their robust natural language processing capabilities. While they share this common architecture, they differ significantly in their training methodologies and objectives. BERT, introduced by \cite{Devlin2019}, employs two distinct pretraining objectives: Masked Language Modeling (MLM) and Next Sentence Prediction (NSP). In the MLM task, certain words within a sentence are intentionally masked, and the model's goal is to predict these masked words using the context provided by the surrounding words. Similarly, in the NSP task, the model is presented with pairs of sentences and must determine whether the second sentence logically follows the first or if it is a random, unrelated sentence. In contrast, RoBERTa\cite{Liu2019} focuses solely on the MLM objective, excluding the NSP task altogether, which has been shown to perform better in various benchmarks. 

For our study. we pretrained single BERT \cite{Devlin2019} variant comprising 110 million parameters using the tokenizer of vocabulary size 30,522. Similarly, we also pretrained the single RoBERTa\cite{Liu2019} variant, also comprising 110 million parameters, using the tokenizer of vocabulary size 50,257. 

In the case of both BERT (Devlin et al., 2019) \cite{Devlin2019} and RoBERTa \cite{Liu2019}, we used a batch size of 256 and trained for 400k steps. We chose the Adam optimizer \cite{Kingma2014} with a learning rate of $1 \times 10^{-4}$, $\beta_1 = 0.9$, $\beta_2 = 0.999$, and included an L2 weight decay of 0.01. We also implemented a learning rate warmup for the first 10,000 steps, followed by a linear decay to ensure smooth training. To improve generalization, we set a dropout probability of 0.1 on all layers. For activation, we used the Gaussian Error Linear Unit (GELU) \cite{Hendrycks2016} function.

\begin{figure}[t]
  \includegraphics[width=\columnwidth]{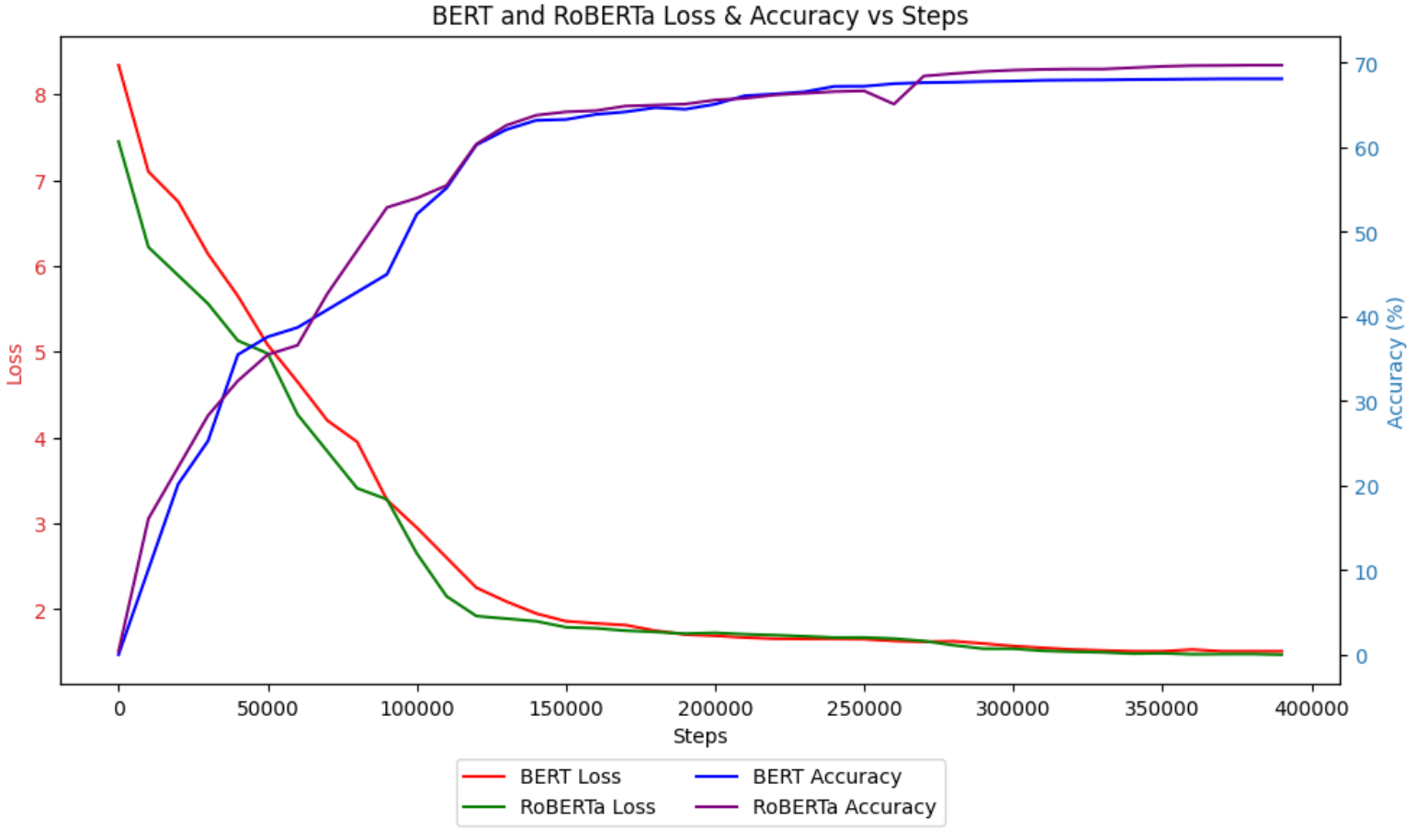}
  \caption{Loss and accuracy of the BERT and RoBERTA model compared with steps}
  \label{fig:experiments_bert_roberts}
\end{figure}

The training loss and accuracy trends for BERT and RoBERTa models are illustrated in Figure \ref{fig:experiments_bert_roberts}. For BERT, the training loss starts at 8.34 and gradually decreases to 1.51 after 400k steps, while accuracy improves from 3\% to 68.11\%. In comparison, RoBERTa begins with a training loss of 7.45, which steadily drops to 1.47 by 400k steps, achieving a slightly higher accuracy of 69.72\%. The figure underscores RoBERTa's faster convergence and marginally better performance than BERT.

\section{GPT-2}
\begin{figure}[t]
  \includegraphics[width=\columnwidth]{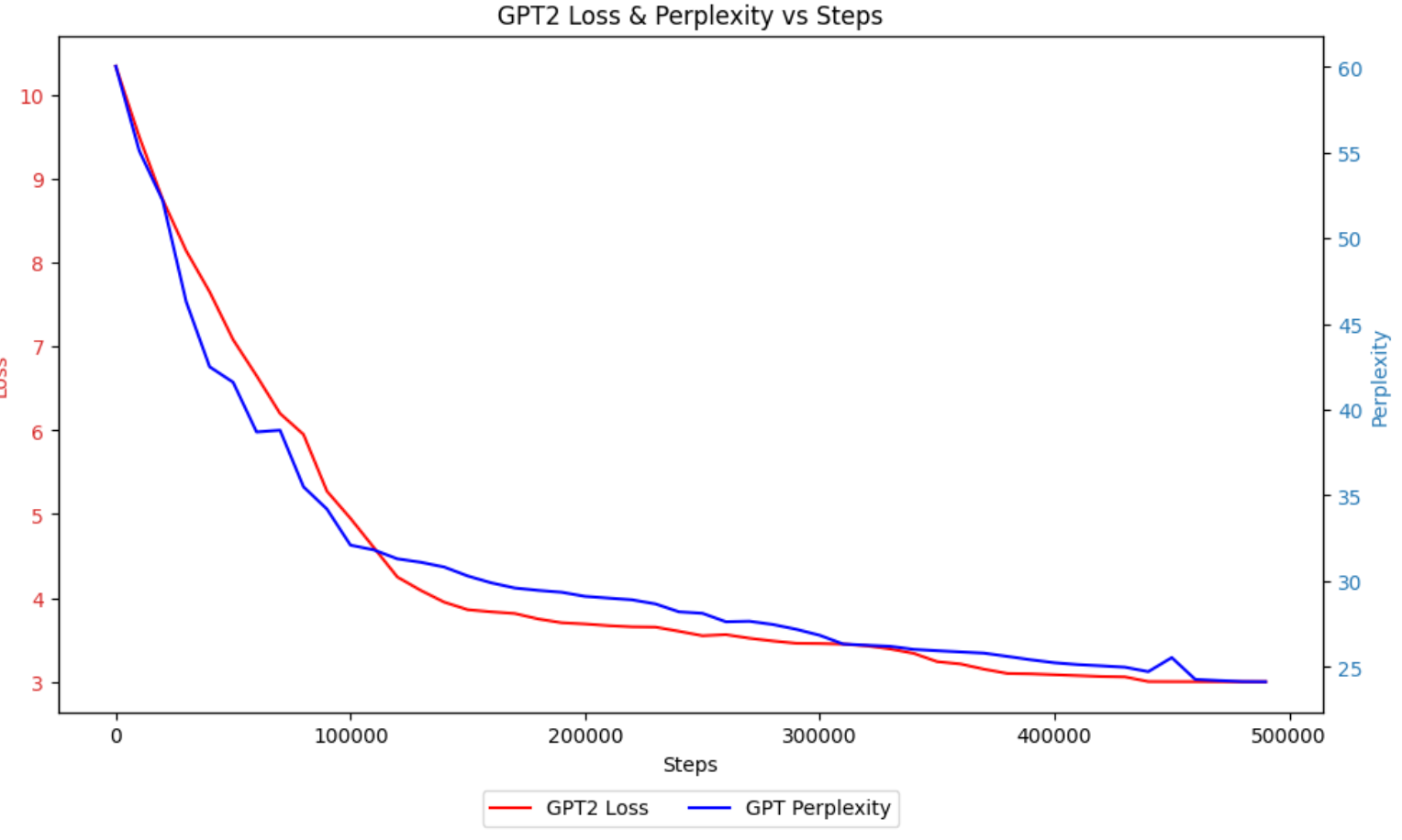}
  \caption{Loss and Perplexity of the GPT2 model compared with steps}
  \label{fig:experiments}
\end{figure}

In the case of GPT-2 \cite{Radford2019}, we pretrained the 124M parameter model using the causal language modeling (CLM) objective, as described in the original GPT-2 paper \cite{Radford2019}. CLM, as outlined in the original GPT-2 paper \cite{Radford2019}, is designed to predict the next word in a sequence given the preceding context. Unlike BERT's masked language modeling, which predicts masked words from the surrounding context, CLM operates in a left-to-right manner. This means that the model generates text sequentially, using only the preceding words to predict the next word in the sequence. We used a batch size of 256 and trained for 500k steps. We used the Adam optimizer with a learning rate of $1 \times 10^{-4}$, $\beta_1 = 0.9$, $\beta_2 = 0.98$, and an L2 weight decay of 0.01. Similar to BERT \cite{Devlin2019} and RoBERTa \cite{Liu2019}, we implemented a learning rate warmup over the first 10,000 steps, followed by a linear decay. To regularize the model, we set the dropout probability to 0.1 across all layers. This model was also trained using the GELU activation function \cite{Hendrycks2016}. Furthermore, we performed instruction tuning on the pretrained model using supervised fine-tuning. We used a batch size of 16, the Adam optimizer with a learning rate of $1 \times 10^{-4}$, and an attention dropout probability of 0.01. 
The training loss and perplexity trends for GPT-2 are shown in Figure \ref{fig:experiments}. Initially, the training loss starts at 10.34 and steadily decreases to 3.001 after 500k steps. Similarly, perplexity drops from 60.03 to 24.13 over the same period. The figure illustrates GPT-2's significant improvement in model performance, with both loss and perplexity showing a consistent downward trend, reflecting the model's enhanced ability to predict the next token with greater accuracy as training progresses.

\section{Evaluation}
We conducted a thorough evaluation across several NLP tasks. Our evaluation includes both Natural Language Understanding (NLU) for encoder models and Natural Language Generation (NLG) tasks for decoder models.

 \subsection{Evaluating BERT \& RoBERTa}
For the evaluation of encoder-based models (BERT \& RoBERTa) \cite{Devlin2019, Liu2019}, we used the Nepali Language Evaluation Benchmark, or Nep-gLUE \cite{Timilsina2022}. It consists of four tasks, including Named Entity Recognition (NER), Part-of-Speech (POS) Tagging, text classification, and categorical pair similarity. We used a batch size of 32 and fine-tuned for 3-10 epochs with multiple learning rates (5e-5, 4e-5, 3e-5, 2e-5, and 1e-5) over the data for all Nep-gLUE tasks. For each task, we selected the best-performing model on the test set.

Our models outperformed all existing models across all tasks, scoring 95.60 on Nep-gLUE \cite{Timilsina2022} benchmark, a result that can be primarily attributed to the large and diverse training corpus we used. The scale of the data allowed the models to generalize better and capture a broader range of linguistic patterns, leading to improved performance.

\begin{table*}[h!]
  \centering
  \resizebox{\textwidth}{!}{%
    \begin{tabular}{c|c|c|c|c|c|c}
      \hline
      \textbf{Model} & \textbf{PARAMS} & \textbf{NER} & \textbf{POS} & \textbf{CC} & \textbf{CPS} & \textbf{Nep-GLUE Score} \\
      \hline
      multilingual BERT \cite{Devlin2019} & 172M & 85.45 & 94.65 & 91.08 & 93.60 & 91.19 \\
      XLM-Rbase \cite{Conneau2020} & 270M & 87.59 & 94.88 & 92.33 & 93.65 & 92.11 \\
      NepBERT \cite{Pudasaini2023}  & 110M & 79.12 & 90.63 & 90.98 & 91.05 & 87.94 \\
      NepaliBERT \cite{rajan2021github}  & 110M & 82.45 & 91.67 & 90.10 & 89.46 & 88.42 \\
      NepBERTa \cite{Timilsina2022}& 110M & 91.09 & 95.56 & 93.13 & 94.42 & 93.55 \\
      BERT (Ours) & 110M & 93.57 & 96.94 & 94.47 & 95.72 & 95.18 \\
      RoBERTa (\textbf{Ours}) & 125M & \textbf{93.74} & \textbf{97.52} & \textbf{94.68} & \textbf{96.49} & \textbf{95.60} \\
      \hline
    \end{tabular}
  }
  \caption{Nep-gLUE Test Result}
  \label{tab:bert_roberta_evaluation}
\end{table*}

\subsection{Evaluating GPT-2}
There were no existing benchmarks for NLG tasks, so we used abstractive summarization for the evaluation of GPT-2 \cite{Radford2019}. We used a publicly available summarization dataset \cite{Bhandari2024} and fine-tuned both the GPT-2 model and GPT-2 Instruct model. The dataset consists of 7,258 data points, where we used 5,806 (80\%) data points for training and the remaining 1,452 (20\%) data points for evaluation. For finetuning, we used a batch size of 8 and trained for 3,000 steps. We used the ROUGE score (Lin, 2004) \cite{Lin2004} as our evaluation metric.

One of the things we observed was that the model tends to hallucinate when given very long contexts, and it did not perform well on long inputs, typically those exceeding 400 tokens. A key reason for this behavior can be traced to the model's training. The model was originally trained on sequences of 512 tokens, which limits its ability to handle longer sequences effectively, resulting in average ROGUE scores.

\begin{table*}[h!]
  \centering
  \normalsize 
  \scalebox{0.9}{ 
    \begin{tabular}{c|c|c|c|c}
      \hline
      \textbf{Model} & \textbf{PARAMS} & \textbf{ROUGE-1} & \textbf{ROUGE-2} & \textbf{ROUGE-L} \\
      \hline
      distilgpt-nepali \cite{Maskey2022} & 88.2M & 10.16 & 8.63 & 9.19  \\
      GPT-2 (Ours) & 124M & 19.66 & 14.51 & 16.84  \\
      GPT-2-Instruct (\textbf{Ours}) & 124M & \textbf{20.42} & \textbf{15.89} & \textbf{17.76} \\
      \hline
    \end{tabular}
  }
  \caption{Performance comparision on summarization task}
  \label{tab:gpt2_evaluation}
\end{table*}

\section{Results}
We evaluated our pretrained Nepali language models on various Natural Language Processing tasks, comparing their performance with existing models. The results of the evaluation are summarized in table \ref{tab:bert_roberta_evaluation} and table \ref{tab:gpt2_evaluation}.

For BERT and RoBERTa in table \ref{tab:bert_roberta_evaluation}, we used the NepGLUE benchmark and evaluated models, against existing monolingual and multilingual models. Both of our models outperformed the previous state-of-the-art, with RoBERTa achieving the highest overall NepGLUE score of 95.60. In particular, our models demonstrated superior performance on every task, reflecting the effectiveness of our models.

In the summarization task in table \ref{tab:gpt2_evaluation}, we compared our GPT-2 models, including an instruction-tuned variant with existing \cite{Maskey2022} model. Our GPT-2 models demonstrated substantial improvements across all ROUGE metrics, with the GPT-2-Instruct model achieving the highest scores of 20.42 (ROUGE-1), 15.89 (ROUGE-2), and 17.76 (ROUGE-L).

\section{Conclusion}
Our study reports significant progress in the field of Natural Language Processing (NLP) for the Nepali language, achieved through the development and evaluation of pre-trained large language models. Our key contributions include the development of by far the largest monolingual corpus for Nepali language and the pretraining of RoBERTa and BERT variants, as well as the introduction of the first GPT-2 model specifically designed for Nepali.

Extensive evaluations conducted on the NepGLUE benchmark and abstractive summarization tasks reveal that our models outperform existing state-of-the-art methods, demonstrating substantial improvements across a range of NLP tasks. By addressing both encoder and decoder architectures, our research emphasizes the potential for optimizing language models tailored to low-resource languages. Our findings not only contribute to the existing body of knowledge but also lay the groundwork for future research and applications in low-resource settings. We anticipate that these insights and benchmarks will inspire further innovations in the field, ultimately resulting in more effective and inclusive NLP research.

\section{Acknowledgement}
We extend our sincere gratitude to Google’s TPU Research Cloud program for granting us free and unlimited access to TPU v4-8 for 30 days and School of Engineering, Kathmandu University for providing us with Nvidia GeForce RTX 3090 GPUs. This research would not have been possible without the unwavering support of the TPU Research Cloud team and School of Engineering, Kathmandu University.

\bibliography{custom}

\end{document}